\documentclass[sigconf]{acmart}
\usepackage{xspace}

\AtBeginDocument{%
  \providecommand\BibTeX{{%
    \normalfont B\kern-0.5em{\scshape i\kern-0.25em b}\kern-0.8em\TeX}}}
\usepackage{balance}

\acmConference[HILDA '20]{HILDA '20: ACM SIGMOD Workshop on Human-In-the-Loop Data Analytics}{June 19, 2020}{ Portland, OR}

\renewcommand\footnotetextcopyrightpermission[1]{}

    {\end{list}}
    
\newcommand{\frameme}[1]{
\vspace{1pt}
\noindent\fbox{
  \parbox{0.95\linewidth}{
    \noindent #1
    }
  }
\vspace{2pt}
}

\newcommand{\thesistext}[1]{#1}

\begin{document}
\pagestyle{plain}

\title{Demystifying a Dark Art: Understanding Real-World \\
Machine Learning Model Development}

\author{Angela Lee}
\authornote{Equal Contribution.}
\affiliation{%
  \institution{University of Illinois at Urbana-Champaign}
}
\email{alee107@illinois.edu}

\author{Doris Xin$^*$, Doris Lee$^*$, Aditya Parameswaran}
\affiliation{%
  \institution{University of California, Berkeley}
}
\email{{dorx, dorislee, adityagp}@berkeley.edu}

\renewcommand{\shortauthors}{Lee et al.}

\begin{abstract}
    It is well-known that the process of developing machine learning (ML) workflows is a dark-art; even experts struggle to find an optimal workflow leading to a high accuracy model. Users currently rely on empirical trial-and-error to obtain their own set of battle-tested guidelines to inform their modeling decisions. In this study, we aim to demystify this dark art by understanding how people iterate on ML workflows in practice. We analyze over 475k user-generated workflows on OpenML, an open-source platform for tracking and sharing ML workflows. We find that users often adopt a manual, automated, or mixed approach when iterating on their workflows. We observe that manual approaches result in fewer wasted iterations compared to automated approaches. Yet, automated approaches often involve more preprocessing and hyperparameter options explored, resulting in higher
performance overall---suggesting potential benefits for a human-in-the-loop ML system that appropriately recommends a clever combination of the two strategies.

\end{abstract}

\maketitle

\section{Introduction}
Machine learning (ML) has become a critical component of almost every domain, with users of all different levels of expertise relying on ML models to accomplish specific tasks. Developing an ML workflow can be tedious and time-consuming. Anecdotally, it may take dozens of iterations for a novice to converge on a satisfactory combination of ML model type, hyperparameters, and data preprocessing. Meanwhile, an expert might need only a small number of modifications to their original workflow to achieve comparable performance on the same task.
In such cases, novices can benefit
greatly from learning strategies employed by experts to accelerate the process of developing
effective ML workflows. With a large-scale database of user-generated workflows and evaluation scores, it becomes possible to extract aggregate workflow iteration patterns to enable this transfer of knowledge.

In this work, we present useful insights about ML workflow development behavior of users on the OpenML\cite{Vanschoren2013} platform. OpenML is an open-source, hosted platform for users to upload datasets and run ML workflows on these datasets by calling an API. 
A relatively diverse mix of user skill levels is present on OpenML, from students just getting started with ML to experienced data scientists and ML researchers.
OpenML publishes a database of the user-uploaded datasets as well as the workflow specifications submitted by users and their corresponding executions.
By performing targeted analyses on the most common ML models and preprocessing operators as well as their associated performance, we shed light on common ML workflow design patterns and their general effectiveness. We study trends in iterative ML workflow changes across 295 users, 475,297 runs, and 793 tasks on the OpenML platform; and draw quantitative conclusions from this crowd-sourced dataset, leading to actionable insights that help inform 
the development of future human-in-the-loop ML systems targeting novices and experts alike. 

Our main contributions can be summarized as follows:
\vspace{-3pt}
\begin{itemize}
\item We characterize the frequency and performance of popular ML models and preprocessing operators used by OpenML users. We highlight the impact of the operator combinations and discuss their implications on general user awareness (or lack thereof) of particular ML concepts (Section~\ref{sec:run}).
\item We analyze sequences of changes to workflows to extract different styles of ML workflow iteration and shed light on the most common types of changes for each iteration style, the amount of exploration typically performed, and performance gain users are generally able to achieve (Section ~\ref{sec:seq}).
\item We conduct case studies on exemplary instances to understand effective iteration practices. (Section~\ref{sec:case}).
\end{itemize}

\noindent\textbf{Outline}. The rest of this paper is organized as follows. After a discussion of related work in Section ~\ref{sec:relatedwork}, welay out terminology and briefly explain our data processing procedure in Section~\ref{sec:method}. In Section~\ref{sec:run}, we shed light on key characteristics of the workflows commonly designed by OpenML users. We investigate the shape and evolution of workflows across iterations in Section~\ref{sec:seq}, examine case studies in Section~\ref{sec:case}, and present final concluding insights and future work in Section~\ref{sec:conclusion}.

\section{Related Work}
\label{sec:relatedwork}
There is an incredible wealth of knowledge that can inform the design of automated and semi-automated ML (autoML) systems by understanding how people develop machine learning models. Studies through empirical code analysis and qualitative studies offer different lenses into studying human-centered practices in developing ML workflows. 
\par Psallidas et al.\cite{Psallidas2019} analyzed publicly-available computational notebooks and enterprise data science code and pipelines to illustrate growing trends and usage behavior of data science tools. Other studies have employed qualitative, semi-structured interviews to study how different groups of users engage with ML development, including how software engineers~\cite{Amershi2019} and non-experts~\cite{Yang2018} develop ML-based applications, and how ML practitioners iterate on their data in ML development~\cite{Hohman2020}.
\par In this study, we analyze practices and behaviors for users iterating on ML workflows, based on data collected from OpenML~\cite{Vanschoren2013}---an online platform for organizing and sharing ML experiments to encourage open science and collaboration. Data from OpenML has been used extensively for meta-learning~\cite{Vanschoren2018} to recommend optimal workflow configurations, such as
preprocessing~\cite{Bilalli2019, Post2016},
hyperparameters~\cite{VanRijn2018}, and models~\cite{Bilalli2017, Strang2018} for a given dataset and task. Benchmark datasets from OpenML, as well as other similar dataset repositories~\cite{Olson2017, d3m}, have also been used for evaluating AutoML algorithms~\cite{Gijsbers2019, Cava2019, Fusi}. However, these papers focus solely on using the dataset, model and result from each workflow and do not study the iterative behavior of users. 
\par Our research differs from existing work in that we analyze the trace of \emph{user-generated} ML workflow iterations leading to insights such as which stages (preprocessing, model selection, hyperparameter tuning) users focus most of their iterations on, their impact on the effectiveness of
the workflow, and which specific combinations of model and preprocessing operators are the most widely used. Our approach offers a more reliable view of iterative ML development than the previous work using experiments reported in applied ML papers to approximate iterations~\cite{xin2018developers}.
Moreover, our study offers a complementary perspective to existing interview studies by providing empirical (based on code), population-level statistics and insights on how people iterate on machine learning workflows.
\par These insights are valuable for helping AutoML and mixed-initiative-ML system builders understand the trends in the ML development practices of target users. System builders can leverage this information about the ML model and preprocessing usage behaviors of the target population to design tools that cater to the needs and iterative styles of the user base. In particular, learning common user habits would help inform \emph{interactive} machine learning, a space in which it is especially imperative to have a strong understanding of the user, and which existing systems sometimes fail to account for~\cite{Amershi2014}. Therefore, the data-driven insights that we provide are a step towards understanding machine learning strategies and bottlenecks from the masses that can help inform the creation of effective crowd-powered systems.

\newcommand{\mean}{$\mu_t$\xspace}
\newcommand{\seqAUC}{$p_S$\xspace}
\section{Data \& Methodology}
\label{sec:method}
We define important terms used in our study, describe our dataset, and define the metrics for quantifying workflow effectiveness.

\subsection{Terminology}
\begin{figure*}
\centering
  \includegraphics[height=0.9\textheight]{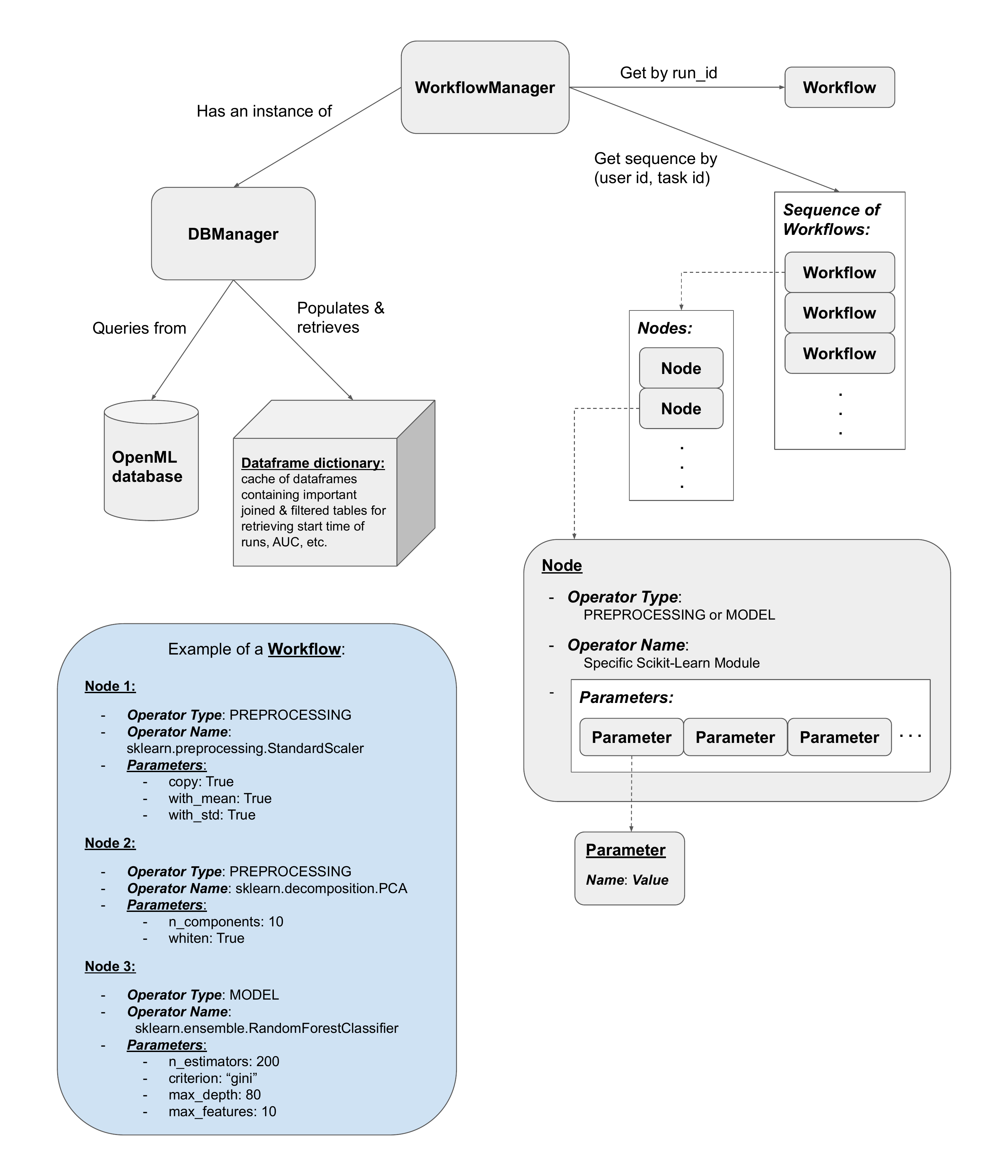}
  \caption{Simplified diagram showing the extraction of workflows from the OpenML database, and an example of what a workflow could look like (blue box).}
  \label{fig:workflowobjschematic}
\end{figure*}

Some terms that will be frequently used throughout this paper are:
\begin{itemize}
    \item \textbf{Task}: A task consists of a dataset along with a machine learning objective, such as clustering or supervised classification.
    \item \textbf{Model}: A model is an implementation of a machine learning algorithm, also known as a classifier or learner. A list of models that are prevalent on OpenML, and therefore will be discussed throughout this paper, is shown in Table~\ref{tab:modelsref}. 
    \item \textbf{Preprocessing}: A preprocessing operator is an implementation of an algorithm that transforms or subsets the data before a model is applied to it. Examples include data cleaning (such as imputation), sampling, feature scaling, or normalization.
    \item \textbf{Workflow}: A workflow is a directed acyclic graph of data \textit{preprocessing} and ML \textit{model} operators with their associated hyperparameters.
    \item \textbf{Run}: A run is a workflow applied to a particular task. Each run is associated with performance measures, such as classification accuracy or Area Under the ROC Curve (AUC).
    \item \textbf{Sequence}: A sequence consists of the time-ordered set of runs from a single user for a particular task.
\end{itemize}
For instance, say a user is working on building an ML \emph{workflow} for a binary classification \emph{task}. The user constructs an initial workflow consisting of a two data \emph{preprocessing} steps---first Standard Scaler, then Principal Component Analysis (PCA) for dimensionality reduction---followed by the application of a Random Forest \emph{model}. The user tests the workflow on the dataset for the task, views the AUC score, then tweaks a few hyperparameters to try to increase the AUC, running and evaluating the workflow after each adjustment. After iterating on the workflow ten times, the user has produced a \emph{sequence} of ten \emph{runs}. A more detailed illustration of a workflow is shown in the blue box in Figure~\ref{fig:workflowobjschematic}.

\begin{table*}
\centering
\small
  \begin{tabular}{|p{2.5cm}|p{13.5cm}|}
    \toprule
     Scikit-Learn Model & Brief Description\\ 
    \hline\hline
     Decision Tree & A tree structure in which each internal node splits into children nodes based on the outcome of a ``test'' on an attribute, and each leaf node contains a class label. From root to leaf, each path represents a classification rule.\\
     \hline
     Random Forest & An ensemble of Decision Tree classifiers trained on sub-samples of the dataset and averaged to improve predictive accuracy and control over-fitting.\\
     \hline
     Extra Tree & A randomized tree classifier in which the best split for the samples of a node is chosen from a set of random splits for each of the randomly selected features.\\
     \hline
     Extra Trees & An ensemble of Extra Tree classifiers.\\
     \hline
     C-Support Vector Classification (SVC) & A Support Vector Machine (SVM) with regularization parameter C. The general idea of an SVM is to find hyperplanes that best divide the dataset into classes.\\
     \hline
     Linear SVC & Similar to SVC but with a linear kernel and implemented in a way that provides more flexibility in the choice of penalties and loss functions, allowing it to scale better with high numbers of samples.\\
     \hline
     Logistic Regression & Models the probability of the default class (in a binary classification problem) using the logistic function.\\
     \hline
     KNeighbors & A classifier implementing the k-nearest neighbors voting rule, in which the labels of the k training samples closest in distance to the new sample are used to predict the label of that new sample.\\
     \hline
     Gaussian Naive Bayes (GaussianNB) & An application of Bayes' theorem (assuming conditional independence between each pair of features given the class label), with the additional assumption that the likelihood of the features is Gaussian.\\
     \hline
     Gaussian Process & A probabilistic model where a Gaussian Process prior is placed on a latent function, which is then passed through a link function to get the probabilistic classification.\\
     \hline
     Multi-layer Perceptron (MLP) & A neural network model that optimizes the log-loss function using quasi-Newton methods or stochastic gradient descent.\\
     \hline
     AdaBoost & A meta-estimator that starts by fitting a base classifier on the dataset, then increases the weights of erroneously classified instances to train subsequent copies of the classifier (with the adjusted weighting) to focus on the more difficult cases. These weak learners contribute to the final strong learner based on their performance.\\
     \hline
     Gradient Boosting & Similar to AdaBoost, but instead of iteratively training on the original dataset (with new sample weights), weak learners are trained on only the remaining errors. The strong learner is then computed through a gradient descent optimization process.\\
     \hline
     XGBoost (XGB) & An implementation of gradient boosted decision trees designed with system and algorithmic optimizations for speed and performance.\\
     \hline
     Voting & Majority rule voting for the class labels predicted by a list of inputted estimators.\\
  \bottomrule
\end{tabular}
  \caption{Brief description of the most commonly used Scikit-Learn\cite{scikit-learn} (sklearn) models on OpenML. Note that the ``Classifier'' suffix in models such as ``Voting Classifier'' are omitted throughout this paper so that the eye can be drawn towards the key points rather than having tables and plots cluttered with the same suffix in every other model name.}
  \label{tab:modelsref}
\end{table*}

\subsection{The Dataset}
Our dataset is derived from a snapshot of the OpenML database from December 4, 2019. The OpenML database contains dozens of tables holding information about datasets and runs uploaded by users, math functions for generating metrics such as evaluation scores for the runs, meta-data about dataset features, and the components of the algorithms that utilized in each run. We extract the workflow-specific information through a WorkflowManager class that we designed as shown in Figure~\ref{fig:workflowobjschematic}. The OpenML data (handled by DBManager) is parsed into workflow objects, which consist of a list of nodes corresponding to each machine learning operator used in the workflow. A set of parameters is associated with each node as well.

We focus on a specific subset of the runs on OpenML that:
\begin{itemize}
    \item \emph{Were uploaded by users who are not OpenML developers}. We filtered out runs uploaded by the core team and key contributors\footnote{listed on \url{https://www.openml.org/contact}} or bots (users whose names contained ``bot''). Our motivation for doing this was to focus on realistic human user behavior so that our insights would be grounded on data pertaining to a more accurate representation of the general ML community. Removing developers and bots left us with 6.1\% of the total number of runs on OpenML. 
    \item \emph{Use the Scikit-Learn package~\cite{scikit-learn}}. Focusing on a single package made it more convenient to parse machine learning operator names and structures (rather than needing to build a separate parser for each library). Furthermore, Scikit-Learn is the most popular machine learning package used on OpenML, with usage in 79\% of the non-developer/non-bot runs. Therefore, even when we limited our processing to only Scikit-Learn workflows, we were still analyzing a representative population.
    \item \emph{Have an associated AUC score associated with them}. By filtering out runs with missing AUC scores, we eliminated missing values when calculating workflow performance metrics. This final subset contains 475,297 runs (4.8\% of the total number of runs). 
\end{itemize}

\noindent\textbf{Limitations:} While the OpenML dataset provides a valuable probe into how users iterate on ML models, the typical OpenML user may not be representative of general ML practitioners. The OpenML traces provide a limited view of the user’s motivation and thought process behind their iteration choices. Our following analysis is intended to present a formative picture of how people iterate in ML development. Future studies with larger sample sizes and more representative samples are required to generalize these findings.

\subsection{Workflow Effectiveness Metrics}
Due to the variability in the range of evaluation metric values across tasks, raw AUC values of runs are not directly comparable across tasks. For instance, an AUC of 0.75 may be below average for a given task, but it may be in the top tenth percentile for another task that is much more difficult.
Therefore, we account for the difficulty of each task by measuring each run's \textit{relative AUC}, defined as the difference between the run's raw AUC and the average AUC of all the runs for the same task.
More formally, let $t$ be an arbitrary task, and $R_t$ be the set of runs that were done on that task. For $r \in R_t$ with a raw AUC of $a_r$, its relative AUC $p_r$ is calculated as:
$$p_r = a_r - {\frac{1}{|R_t|} \sum_{k \in R_t} a_k}$$

We measure the performance of a sequence $S$ by the \textit{relative maximum sequence AUC}, 
\begin{equation*}
    p_S = \max_{r \in S}{a_r} - {\frac{1}{|\mathcal{T}_t|} \sum_{Q \in \mathcal{T}_t} \max_{q \in Q}{a_q}}
\end{equation*}
, where $\mathcal{T}_t$ is the set of all sequences for task $t$. 
In other words, $p_S$ is the difference between the maximum AUC in the sequence $S$ and the mean of the maximum AUCs of all sequences for task $t$. We use maximum AUC per sequence to capture the fact that the best-performing workflow out of all attempts is the final one that the user adopts.
The $p_S$ metric is used as one of the measures of performance in Section~\ref{sec:seq}, while $p_r$ is used in Section~\ref{sec:run} on run-level insights.

\section{Run-Level Insights}
\label{sec:run}
\begin{figure}
\centering
  \includegraphics[width=0.45\textwidth]{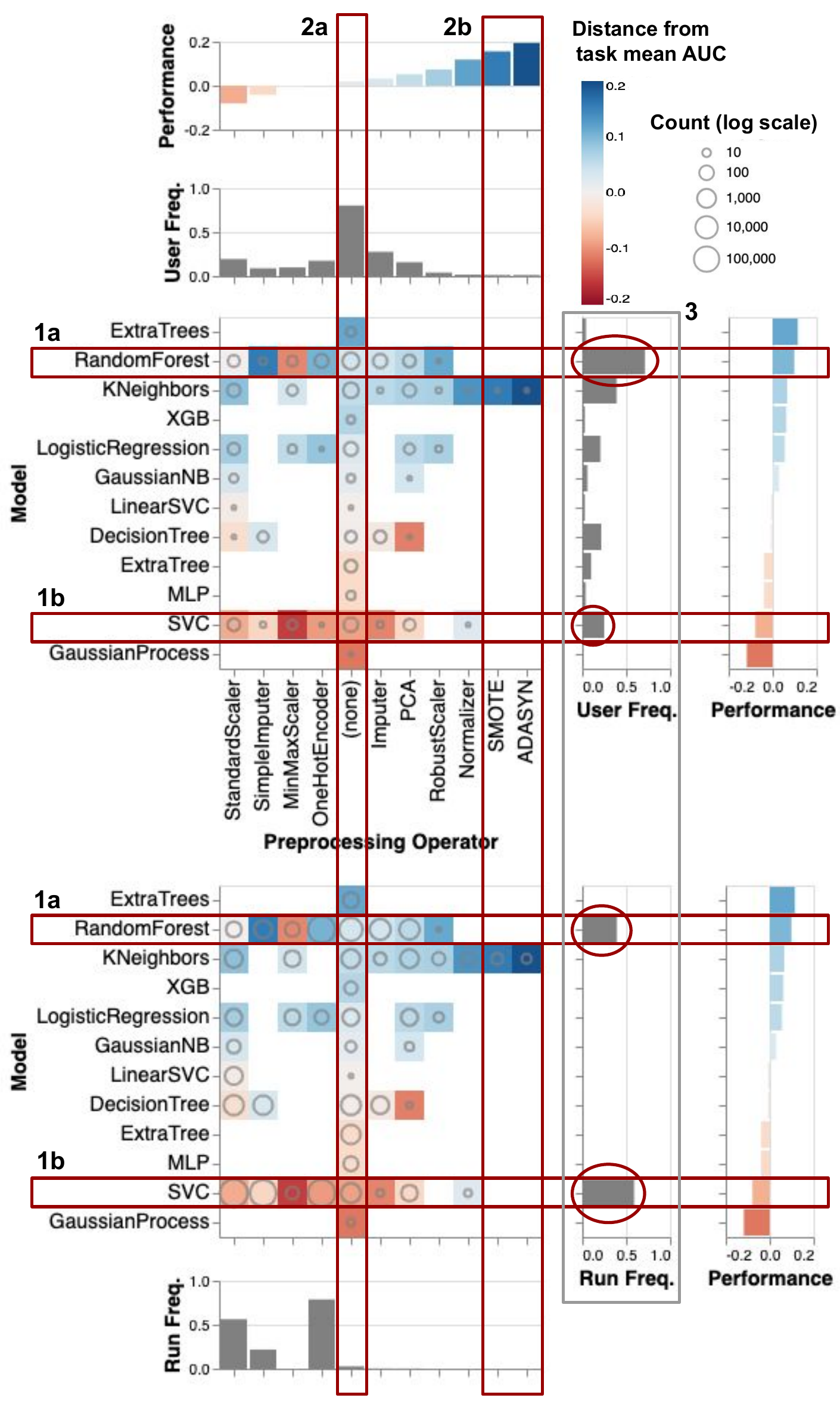}
  \caption{Frequency and performance of the most common (Scikit-Learn model, preprocessing) combinations on OpenML. The top heatmap displays frequency in terms of the user count, while the bottom shows run count.}
  \label{fig:freq_perf_heatmap}
\end{figure}

To better understand how users develop ML workflows, we first sought to understand: \textit{What are the most prevalent ML operators in practice? In what combinations are these operators used together and when do they lead to better performance? Are there any less prevalent combinations that lead to high performance, indicative of ``tricks of the trade'' that only experts seem to know about?}

\par Figure~\ref{fig:freq_perf_heatmap} shows the most common model and preprocessing combinations used by more than four users on OpenML. Certain models (AdaBoost Classifier, Bagging Classifier, Dummy Classifier, Gradient Boosting Classifier, Grid Search CV, Randomized Search CV, and Voting Classifier) were intentionally excluded from the plot. Dummy Classifier was excluded because it is known to be used for comparisons rather than for solving real problems~\cite{scikit-learn}. The others were excluded because they are wrappers that can take in one or more base estimators, and the most frequently used of these base estimators are already shown in the plot.

The color of the square for each combination indicates the performance of the combination, measured by $p_r$, averaged over all runs that include the combination. The size of each circle is scaled by the number of users (Figure~\ref{fig:freq_perf_heatmap} top) and runs (Figure~\ref{fig:freq_perf_heatmap} bottom) that contain the specific combinations. Both the rows and columns are sorted based on their performance across all of the specified operators. 
In other words, the best-performing combinations are located at the top-right corner of the chart, whereas the worst-performing combinations are on the bottom-left.

Histograms showing the marginal distribution of frequency and performance for each operator are displayed on the sides. For instance, the uppermost user frequency histogram shows the average user count for each preprocessing operator, averaged across the models that were used in combination with it, normalized by the total number of users.

We highlight various insights from Figure \ref{fig:freq_perf_heatmap}, with the enumerated points corresponding to the enumerated boxed regions in the figure.

\begin{enumerate}
    \item\textbf{Performance of Specific Combinations:} Some combinations consistently yield high relative AUC when compared to other combinations used for the same ML task. For instance, there is a clear difference in performance between Random Forest (RF) (1a), the model used by the most OpenML users, and SVC (C-Support Vector Classification) (1b), the model used in the most runs. On average, the relative AUC of runs that include Random Forest is $p_r=9.89\%$. RF works especially well with Simple Imputer as an added preprocessing step (for replacing missing values), achieving $p_r = 15.57\%$ on average. On the other hand, users tend to perform worse on average ($p_r=-7.84\%$) when using SVC models for OpenML tasks.
    
    \item\textbf{Effect of Preprocessing:} Around 81\% of users have run ML models on a dataset without performing any data preprocessing beforehand (2a). This could be attributed to the fact that many of the datasets on OpenML are already in a relatively clean, preprocessed state. However, it is evident from Figure \ref{fig:freq_perf_heatmap} that users can often still achieve higher performance by including some form of dimensionality reduction, feature scaling and transformations, or data sampling, indicating how \emph{preprocessing is often an overlooked but important aspect in ML development}. 
    For instance, ADASYN and SMOTE~\cite{AdasynSmote}, which are preprocessing strategies for over-sampling data, are relatively infrequent among OpenML users (2b). However, when combined with K-Nearest Neighbors (KNeighbors) models, users are able to consistently achieve higher AUC scores (by around $p_r=19.4\%$ for ADASYN and $p_r=15.6\%$ for SMOTE). This suggests that when working with imbalanced datasets, certain techniques such as over-sampling are extremely valuable in achieving high classification performance, and that there needs to be increased awareness of such data preprocessing methods and/or when to use them.

    \item\textbf{Frequency of Specific Combinations:} The frequency distributions for both models and preprocessing operators vary depending on whether we look at the number of runs or the number of users (shown in box 3). For example, the most popular model measured by the number of unique users is RF (used by 70.9\% of plotted users and 68.2\% of all users\footnote{We refer to ``all users'' and ``all runs'' out of the complete set of runs containing Scikit-Learn operators and set of uploaders for those runs.}), 
    followed by KNeighbors (38.9\% of plotted users and 37.4\% of all users)
    with SVC in third place (24.8\% of plotted users and 24.2\% of all users). 
    However, when determining frequency from the perspective of the number of runs that included the model, 
    SVC makes up 59.0\% of plotted runs and 47.4\% of all runs, 
    while RF accounts for 39.4\% of plotted runs and 32.3\% of all runs. 
    This suggests that there are variations in the iteration behavior across different users: some focus on tuning the same model for many iterations, while others experiment with several different models. We explore these trends and provide insights on their effectiveness in section~\ref{sec:seq}.
\end{enumerate}

\noindent\textbf{Application of Run-Level Insights:}
Analyses such as those highlighted from (1)-(3) not only shed light on the usefulness of particular ML operators, but can also be used to validate whether users' existing practices aligns with \textit{conventional wisdom} in the form of guidelines from the ML community or whether there is a gap in adopting these guidelines. As an example of this application, we examine a particular case study of how users select which models to use for handling large datasets.

First of all, we observe that for large datasets, users do indeed focus on specific models in a different distribution than the overall trends shown in Figure~\ref{fig:freq_perf_heatmap}.  Table~\ref{tab:bigdatasets} shows the model frequencies for only the runs on large datasets (with greater than 110,313 instances---the mean across all the datasets that Scikit-Learn users constructed workflows for). 
\begin{table}
\centering
  \begin{tabular}{cccc}
    \toprule
     Model & Run Freq & User Freq & Avg $p_r$\\ 
    \hline\hline
     DecisionTree & \textbf{65.64}\% & 6.06\% & 0.57\%\\ 
     KNeighbors & 19.02\% & \textbf{60.61}\% & 0.48\% \\ 
     RandomForest & 9.82\% & 18.18\% & 3.1\% \\ 
  \bottomrule
\end{tabular}
  \caption{Frequency and average distance from task mean AUC of the three most commonly used models for large datasets.}  \label{tab:bigdatasets}
\end{table}

According to conventional guidelines~\cite{Wujek}, Decision Trees and ensemble methods like RF are well-suited for medium to large datasets, while KNeighbors works well for small to medium datasets. Although Decision Tree makes up 65.64\% of the runs, it is used by only 6\% of the users, as shown in Table~\ref{tab:bigdatasets}. Instead, $60.61\%$ of the users opted for KNeighbors for the largest datasets on OpenML. However, KNeighbors resulted in the lowest performance, (average $p_r=0.48\%$), compared to average $p_r=3.1\%$ for RF and 0.57\% for Decision Tree. While the performance validates the efficacy of the guidelines, the usage statistics reveal that most users fail to follow these guidelines.

Insights drawn from empirical run-level data as demonstrated in this section can inform the ML community about the prevalence of different operators and whether or not users are able to effectively use them. By knowing which models and preprocessing operators are most commonly used in practice, ML system designers can learn which algorithms and tools users are already aware of and able to utilize successfully, as well as which ones people are less likely to use yet have high potential for large performance improvements. A human-in-the-loop ML system~\cite{lee2019human} could surface these lesser-known but high-potential operators to educate the users and bridge the gap between the system's capabilities and the user's knowledge on specific capabilities. 

In this section, we have explored various facets of common workflow trends in individuals runs. But to better understand the process of how users are constructing their workflows and iterating on them, we must shift to the wider view of looking at \emph{sequences} of runs for each user and task as a single entity, rather than each run as its own entity. In the next section, we delve into analyzing sequences to gain a perspective on the impact of different iterative changes on the performance of the workflows.

\section{Sequence-Level Insights}
\label{sec:seq}
Users have a wide range of ML workflow development styles---some use a more manual approach where they run a model, look at the results, make a change or two to address the issue, and then repeat the process. Others may choose a more automated technique, e.g., looping through a set of pre-determined values for certain hyperparameters of a model. In the \textit{manual} case, the human remains \textit{in the loop}, while in the \textit{automated} case, the user has already set a search space a-priori and the changes to the workflow at each iteration are independent of the previous iteration's result. Others use a \textit{mixed} sometimes-manual, sometimes-automated strategy.

To classify a sequence as manual, automated, and mixed, we introduce the following metrics.
\begin{itemize}
    \item Interval ($\Delta t$): difference between start times of consecutive runs
    \item Interval difference ($\Delta^2 t$): difference between consecutive $\Delta t$s
    \item Sequence length ($|S|$): the number of runs in sequence $S$
\end{itemize}
Based on these metrics, we categorize each sequence $S$ as follows:
\begin{itemize}
    \item \textbf{Manual}: ($|S| \leq 2$, OR $\Delta t > 10$ minutes for $\geq 50$\% of the runs in the sequence, OR $\Delta^2 t > 3$ minutes for $\geq 75$\% of the runs in the sequence), AND $|S| < 300$
    \item \textbf{Automated} (Auto): $|S| > 2$, AND $\Delta t > 10$ minutes for $< 50$\% of the runs in the sequence, AND $\Delta^2 t > 3$ minutes for $\leq 25$\% of the runs in the sequence
    \item \textbf{Mixed}: the remaining sequences not in the two categories above 
\end{itemize}
The thresholds were empirically determined through a process of random sampling and spot-checking two sequences from the manual category that had over 30 iterations and two from the automated category that had under 30 iterations (since these would be the more ambiguous cases than shorter manual sequences and longer auto sequences). After the final thresholds were set, five sequences from each of the categories were randomly sampled and spot-checked to validate the sequence labels.

The motivation behind looking at both $\Delta t$ and $\Delta^2 t$ is that we would expect the actual time difference between run submissions to be at least a couple of minutes if the user was making adjustments manually, and we would expect the change in these $\Delta t$'s to also be non-zero due to the variability in making changes (while constant time differences are often indicative of a loop). Sequence length is also highly indicative of whether or not most of the runs were manual, since it would be unlikely to find hundreds or thousands of manual runs, and indeed the highest number of runs in a manual sequence (after setting the 300 iteration threshold) is 69.
This categorization results in 2181 manual, 208 automated, and 168 mixed sequences. 

Across the three categories, users exert varying amounts of effort (number of iterations and number of model and preprocessing combinations attempted) and focus on different areas of their workflow (choosing the best model, optimizing a hyperparameter, or determining which data preprocessing operation to add). 
Our in-depth analyses of the three categories of sequences reveal the following three major insights on the effectiveness of users iterating with manual and automated iterations:

\begin{itemize}
    \item \textbf{\emph{On Efficiency}: 
    Users who manually iterate on their workflows are typically more efficient but less effective in improving their ML workflow than those who iterate in an automated manner.} Manual users tend to waste fewer iterations searching after reaching their highest-performing workflow than people who do mostly automated iterations, and they achieve the same performance gain in a small fraction of the number of iterations as automated, meaning that manual users are more efficient. However, from the perspective of effectiveness, on average, \textit{automated users reach a higher maximum AUC than manual for the same task}. This implies that there is still merit to an automated, exhaustive search, indicating the potential benefits of AutoML tools.
    \item \textbf{\emph{On Exploration}: 
    In general, manual users cover the same number of different models in their sequences as automated users, but a lower variety of data preprocessing techniques.} Trying a greater number of different model and preprocessing operators results in a positive impact on performance in manual sequences. However, this trend becomes less visible the more automated the sequences become, and this is likely attributed to the increased amount of sub-optimal hyperparameter tuning present in automated sequences.
    \item \textbf{\emph{On Model Tuning}: Manual users iterate more on model selection than hyperparameter tuning or data preprocessing.} When looking at the specific model changes that are most frequently made, we find (from the percentage of the sequence dedicated to each model and the transition probabilities between the models) that most users are able to eventually converge on higher-performing models.
\end{itemize}

\begin{figure}
  \includegraphics[width=\linewidth]{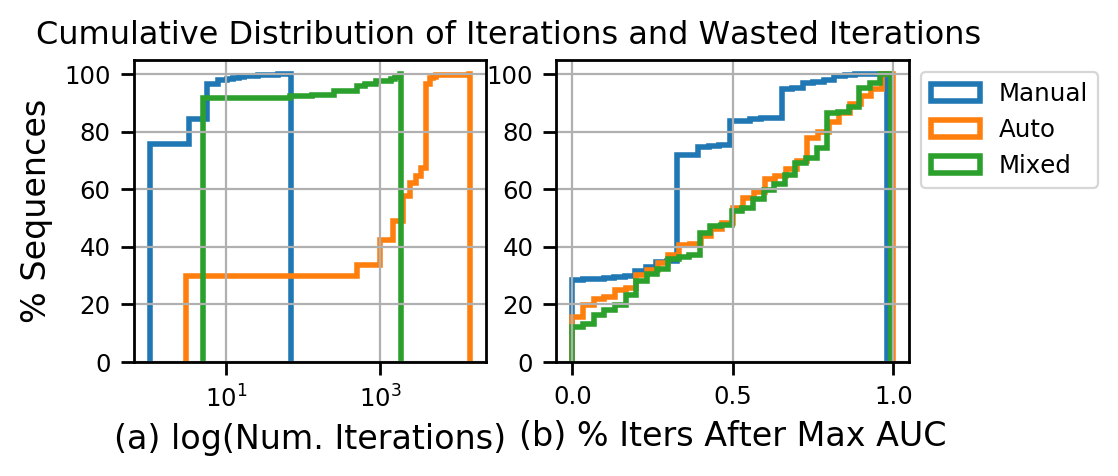}
  \caption{(a) Cumulative distribution of sequence length, (b) \% iterations after reaching maximum AUC.}
  \label{fig:5.1}
\end{figure}

\begin{figure}
  \includegraphics[width=\linewidth]{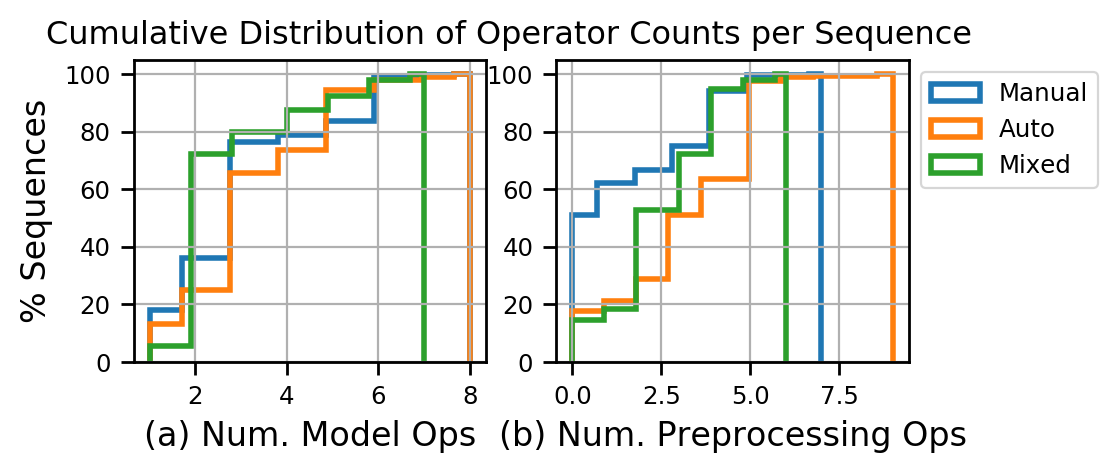}
  \caption{Cumulative distribution of (a) model and (b) preprocessing operators used across all iterations in a sequence.}
  \label{fig:5.2.1}
\end{figure}

\subsection{On Efficiency}
\label{sec:efficiency}
\thesistext{
\frameme{
\textbf{Users iterating with manual sequences are more efficient,
achieving the same performance gain in a small fraction of the number of iterations as automated sequences, while
wasting fewer iterations searching after reaching their highest-performing workflow. However, \textit{users iterating with automated sequences reach a higher $p_S$ than manual for the same task}.}
}
}

\begin{table}
\centering
  \begin{tabular}{cccc}
    \toprule
     Iteration Type & Mean & Median & Standard Deviation\\ 
    \hline\hline
     Manual & 6.63\% & 3.69\% & 8.63\%\\ 
     Mixed & 4.75\% & 0.36\% & 9.29\% \\ 
     Automated & 7.06\% & 1.01\% & 12.66\% \\ 
  \bottomrule
\end{tabular}
\caption{Maximum improvement in AUC from the starting iteration of the sequence.}
\label{tab:deltamaxauc}
\end{table}

Even though manual sequences\footnote{All sequences with a length of 1 were excluded from this and other appropriate analyses in this section to avoid inflation by deltas of 0, counts of 1, or percentages of 100.} are on average $< 0.2\%$ the length of automated sequences, Table~\ref{tab:deltamaxauc} shows that the maximum increase in AUC is equal for manual and automated (Welch's $t=-0.471, p=0.638$). Therefore, manual users more efficient in reaching their best-performing workflow than mixed and auto users, and the performance gain (relative to the performance of the first attempt) of much longer auto sequences is only an insignificant amount higher than manual sequences with far fewer iterations.
Moreover, Figure~\ref{fig:5.1}(b) shows that the manual group has a lower percentage of wasted iterations, i.e., the iterations after $p_S$ has been achieved, than the other two groups (manual: $\mu=31\%$; mixed: $\mu=49\%$; auto: $\mu=47\%$). This means that on average, manual users waste 1 iteration, mixed users waste 41 iterations, and auto users waste 1026 iterations.

However, automated and mixed sequences result in a 3\% higher $p_S$ compared to manual sequences for the same task (Welch's $t=-12.96, p<0.05$). This is likely due to a greater coverage of search space from the mixed and automated sequences compared to manual.
There is a delicate balance between iterating in an efficient manner but exploring enough to achieve better results. We now detail our approach to quantitatively estimating a sequence's breadth of exploration.

\subsection{On Exploration}
\thesistext{
\frameme{
\textbf{Users iterating with manual sequences cover the same number of different models as automated sequences, but a lower variety of preprocessing techniques. 
Manual exploration of more model and preprocessing operators results in higher $p_S$. 
However, automated sequences with more combinations of model and preprocessing operators do not necessarily lead to better performance.}
}
}

As shown in Figure~\ref{fig:5.2.1}, manual sequences explore a similar number of models as mixed and automated sequences (manual: $\mu=3.08, \sigma=1.63$; mixed: $\mu=2.64, \sigma=1.37$; auto: $\mu=3.31, \sigma=1.48$). However, there tends to be less manual exploration on data preprocessing when compared to the mixed and automated groups (manual: $\mu=1.51, \sigma=1.82$; mixed: $\mu=2.49, \sigma=1.44$; auto: $\mu=3.22, \sigma=1.91$). It is interesting to note that automated sequences explore the same number of preprocessing operators as models on average, while manual sequences have a higher tendency to completely leave out data preprocessing from their workflows.

\begin{figure}
  \includegraphics[width=\linewidth]{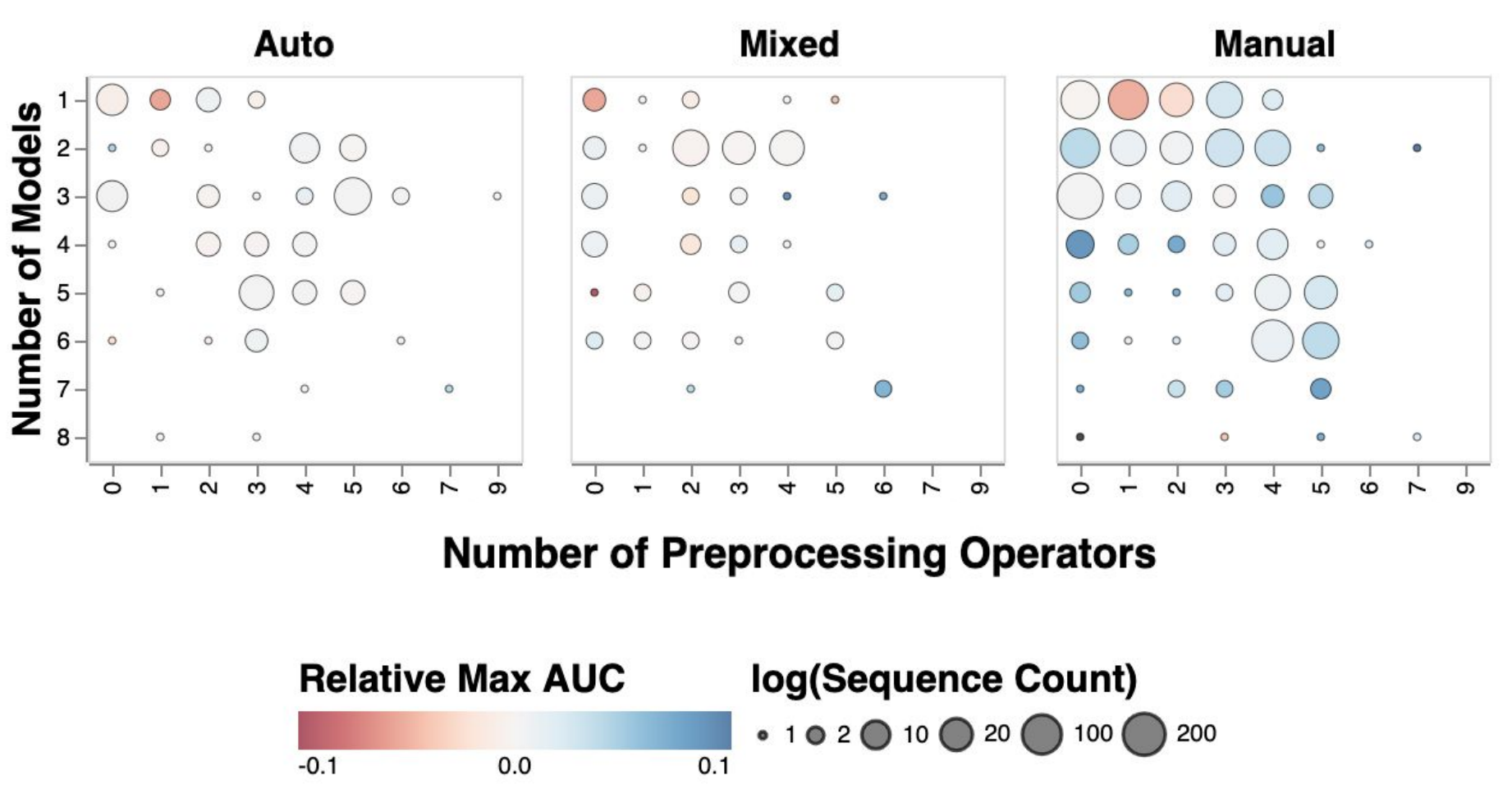}
  \caption{Joint distribution of model and preprocessing operators per sequence. Max AUC is relative to only the sequences within the same category (manual, mixed, or auto).
  }
  \label{fig:5.2.2}
\end{figure}

When examining the number of model and preprocessing operators jointly, as shown in Figure~\ref{fig:5.2.2}, we can see that for manual iterations, trying more combinations leads to better performance, but this is not the case in mixed and auto sequences. 
For each of the combinations that auto sequences explore, they typically perform many more iterations of hyperparameter tuning than manual sequences do for a given combination, evident from the fact that most auto iteration sequences comprise of hyperparameter tuning, as shown in Table~\ref{tab:changetypefreqs}, and that auto sequences are much longer. 

These trends reveal two phenomena: 
1) when a combination is explored using default or rule-of-thumb hyperparameters in just a few iterations, as is the case in typical manual sequences, the performance gap between different combinations is large;
2) when a combination is explored with many hyperparameter tuning iterations, as is the case in typical auto sequences, the performance gap shrinks between different combinations, leading to diminishing returns from exploring more combinations on $p_S$ improvement.
In other words, auto sequences often waste most iterations on hyperparameter tuning to improve the performance of suboptimal combinations without improving $p_S$, while the potential of a combination can be estimated quickly with just a few iterations. However, there is merit to extensive hyperparameter tuning, evident in the fact that auto sequences achieve a 3\% higher $p_S$ than manual ones as discussed in Section~\ref{sec:efficiency}.
Together, these insights suggest that a hybrid approach, wherein coarse-grained search is first performed over the combinations of preprocessing and model operators, followed by fine-grained search doing hyperparameter tuning on the most promising combinations, would be
both effective and also efficient at finding a high-performing workflow.

\subsection{On Model Tuning}
\frameme{\textbf{While automated sequences are dominated by hyperparameter changes, users iterating with manual sequences focus on model selection and eventually converge on high-performing models.}
}

\begin{table}
\centering
  \begin{tabular}{cccc}
    \toprule
     Change Type & Manual & Mixed & Auto\\ 
    \hline\hline
     Model Operator & \textbf{45.04}\% & 3.95\% & 0.18\%\\ 
     Model Hyperparameter & 28.31\% & \textbf{75.23}\% & \textbf{92.69}\% \\ 
     Preprocessing Operator & 0.53\% & 0.55\% & 0.01\% \\ 
     Preprocessing Hyperparameter & 0.18\% & 0.17\% & 0.04\% \\ 
     Model \& Preprocessing & 21.96\% & 6.60\% & 5.45\% \\ 
     No Change & 3.98\% & 13.49\% & 1.63\% \\ 
  \bottomrule
\end{tabular}
\caption{Percentage of runs for each change type across all runs in manual, mixed, and automated sequences. The ``Model \& Preprocessing'' change type refers to a model operator or hyperparameter change, in conjunction with a preprocessing operator or hyperparameter change.
  }
  \label{tab:changetypefreqs}
\end{table}

\begin{table}
\centering
  \begin{tabular}{cccc}
    \toprule
     Model & Manual & Mixed & Auto\\ 
    \hline\hline
     RandomForest & 40.83\% & 76.90\% & 53.32\%\\ 
     KNeighbors & 37.99\% & 37.74\% & 50.54\% \\ 
     SVC & 28.25\% & 23.73\% & 46.26\% \\
  \bottomrule
\end{tabular}
\caption{Mean percentage of iterations in a sequence that included RandomForest, KNeighbors, and SVC (the three most frequently used models) out of the sequences with at least 3 iterations that used the model.
  }\label{tab:percentiterspermodel}
\end{table}

Since changing the ML model is the most common type of manual workflow change, making up 45.04\% of all manual runs as shown in Table~\ref{tab:changetypefreqs}, we look into which models are abandoned and which are kept from one iteration to the next.
From Table~\ref{tab:percentiterspermodel}, we can see that in all groups (manual, mixed, and auto), whenever SVC is used, it does not get used for the majority of iterations---users tend to not stick with it as much as RF or KNN. For instance, whenever SVC is used in a manual sequence, it accounts for only 28.25\% of the runs, compared to 40.83\% for RF and 37.99\% for KNN. Combined with our findings from Figure~\ref{fig:freq_perf_heatmap} in section~\ref{sec:run} which demonstrate that on average, RF and KNN perform better than SVC, we conclude that users are able to waste less time (percentage of their iterations) on models that do not work as well.

\begin{figure}
  \includegraphics[width=0.9\linewidth]{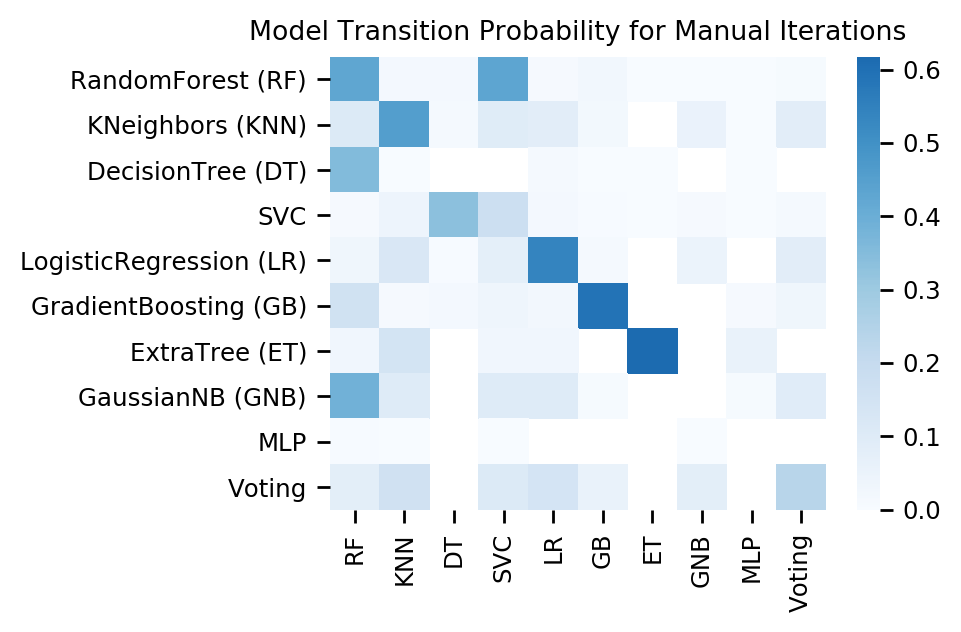}
  \caption{Transition matrix showing the probability of transitioning between different models in consecutive iterations in manual sequences. The model on the y-axis represents the model being used at the current iteration, and the model on the x-axis is the model in the next iteration. The ten most frequent models are displayed and ordered from top to bottom and left to right from highest to lowest user frequency.
  }
  \label{fig:transitionmat}
\end{figure}

\begin{figure}
\centering
  \includegraphics[width=0.9\linewidth]{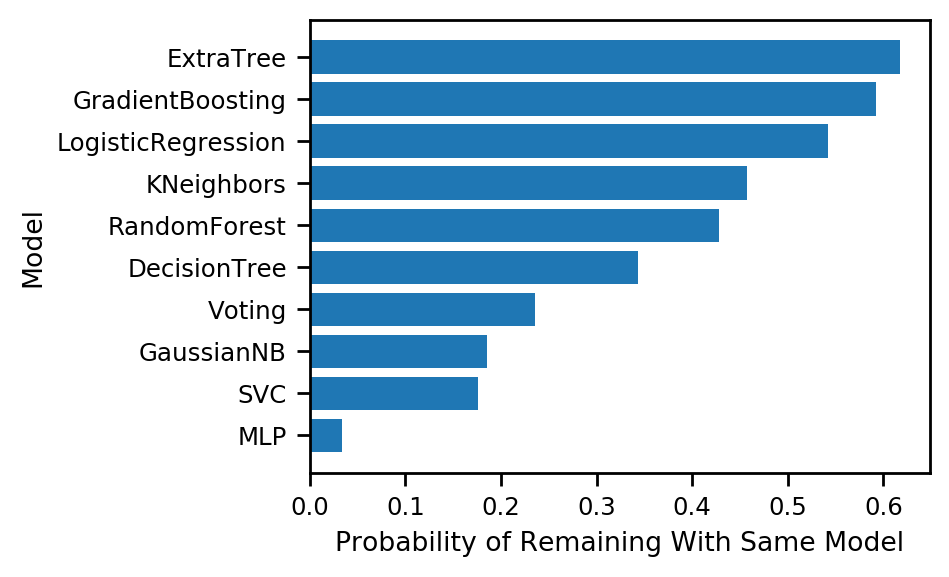}
  \caption{Probability of remaining with the same model in the next iteration (out of manual sequences) for the ten most frequent models.}
  \label{fig:model_nontransition_prob}
\end{figure}

Furthermore, the transition probabilities between different pairs of models in Figure~\ref{fig:transitionmat} reveal users tend to stay with the same model from one iteration to the next, as evidenced by the high probabilities along the diagonal of the matrix. Extra Tree, Gradient Boosting Classifier, and Logistic Regression are the models with the highest probability of not changing between iterations, as displayed in  Figure~\ref{fig:model_nontransition_prob}. Meanwhile, MLP has a notably low probability (3.3\%) of being used again in the following iteration. This presents an interesting area of further investigation to learn why certain models such as MLP are quickly abandoned.

But whenever a switch occurs (from the model listed on the matrix row to the model on the column), the model that is switched to the most is Random Forest, with 26.22\% of all model changes transitioning to RF, as shown by Table~\ref{tab:avgtransitionprob}. Interestingly though, when transitioning away from Random Forest, the most common model in the subsequent iteration is SVC, as can be seen from Figure~\ref{fig:transitionmat}. Then from SVC, Decision Tree is the most probable transition, with an even higher probability than remaining with SVC.

\begin{table}
\centering
  \begin{tabular}{|c|c|}
    \toprule
     Model & Transition Probability to Another Model\\ 
    \hline\hline
     RandomForest & 0.2622\\
     \hline
     SVC & 0.1328\\
     \hline
     KNeighbors & 0.1045\\
     \hline
     LogisticRegression & 0.0594\\
     \hline
     DecisionTree & 0.0579\\
     \hline
     Voting & 0.0446\\
     \hline
     GradientBoosting & 0.0418\\
     \hline
     MLP & 0.0241\\
     \hline
     GaussianNB & 0.0127\\
     \hline
     ExtraTree & 0.0089\\
  \bottomrule
\end{tabular}
\caption{Average probability of transitioning to the model from one of the other ten most frequent models.}\label{tab:avgtransitionprob}
\end{table}

Insights that span multiple workflow iterations, which we have provided in this section, allow us to discover general patterns and better understand common ML development habits. To form an even more detailed view on this subject, we dive into a set of exemplary case studies of individual sequences in the following section.

\section{Case Studies}
\label{sec:case}
So far, we have described insights aggregated across multiple users, reflecting the population-level trends in ML development. In this section, we dive deep into some case studies for concretely visualizing the three categories of iterative behavior, as well as example instances of both \emph{effective} and \emph{ineffective} practices to offer a complementary view on user behavior.

\subsection{On Iterative Behavior}
\begin{figure*}
\centering
  \includegraphics[clip,width=\textwidth]{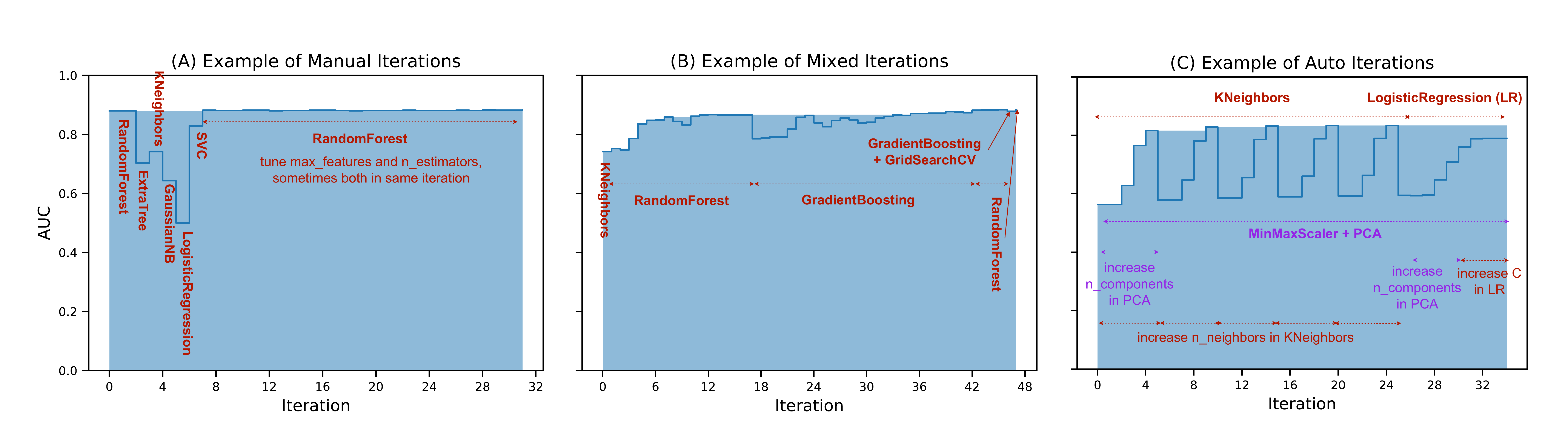}%
\caption{Examples of each of the different categories of sequences: manual (A), mixed (B), and auto (C). The shaded blue area represents the maximum AUC up until that point, the red text describes model changes, and the purple text describes preprocessing changes.}
\label{fig:casestudiestypes}
\end{figure*}

Whereas in the previous section we focused on characterizing auto, mixed, and manual workflows at an aggregate scale, we now present a representative sequence from each of the three categories of iterative styles to provide concrete examples of the differences between these patterns. We randomly sampled from sequences that fall within a small range of total iterations (between 30 and 100), and depict one of each sequence type in Figure~\ref{fig:casestudiestypes}. The examples for auto, mixed, and manual iterations are comparable in length, having between 32 and 48 iterations.

The sequence shown in Figure~\ref{fig:casestudiestypes}(A) was for a supervised classification task on a bioresponse dataset\footnote{https://www.openml.org/d/4134}: predicting whether or not each molecule (row in the dataset) was seen to elicit a biological response to the molecular descriptors that capture some of the characteristics of the molecule (features). 54.23\% of the instances were positive, and there were 1776 already normalized features.
In Figure~\ref{fig:casestudiestypes}(A),
it can be seen that the user started out trying several different ML models before settling on Random Forest because it resulted in the highest AUC out of all the models that the user tried. The user then manually adjusted a few of the hyperparameters for Random Forest for the rest of the iterations, which ended up only raising the maximum AUC by an almost imperceptible amount (0.004). 

The mixed sequence shown in Figure~\ref{fig:casestudiestypes}(B) 
was also done on the same task as the manual example. In the mixed case study, some of the hyperparameters, such as ``learning\_rate'' (from iterations 24 to 29) and ``max\_depth'' (from iterations 30 to 37) were tuned in a loop; meanwhile others were adjusted manually, such as ``n\_estimators'' (from iterations 2 to 6, and again from 10 to 12 and 16, right before switching from Random Forest to Gradient Boosting Classifier). For both Random Forest and Gradient Boosting models, the user was able to have an overall upward trend in AUC over time.

The automated iterations shown in Figure~\ref{fig:casestudiestypes}(C) 
were created for classifying the CIFAR-10 small dataset~\cite{cifar10small}, which contains 2000 images for each of the 10 classes of objects.
This sequence contains a very visible initial subpattern, with no noticeable progress being made overall (outside of the subpattern). The subpattern was likely due to nested loops being run---with the outer loop being the number of neighbors parameter for the KNeighbors model, and the inner loop being the number of components for PCA, one of the preprocessing operators. In this auto sequence, there were only two notable manual iterations: the first was when the model was switched from KNeighbors to Logistic Regression at iteration 26, and the second was at iteration 31 when the user moved from tuning ``n\_components'' in PCA to tuning ``C,'' the inverse regularization strength in Logistic Regression.

\begin{figure*}
\centering
  \includegraphics[clip,width=\textwidth]{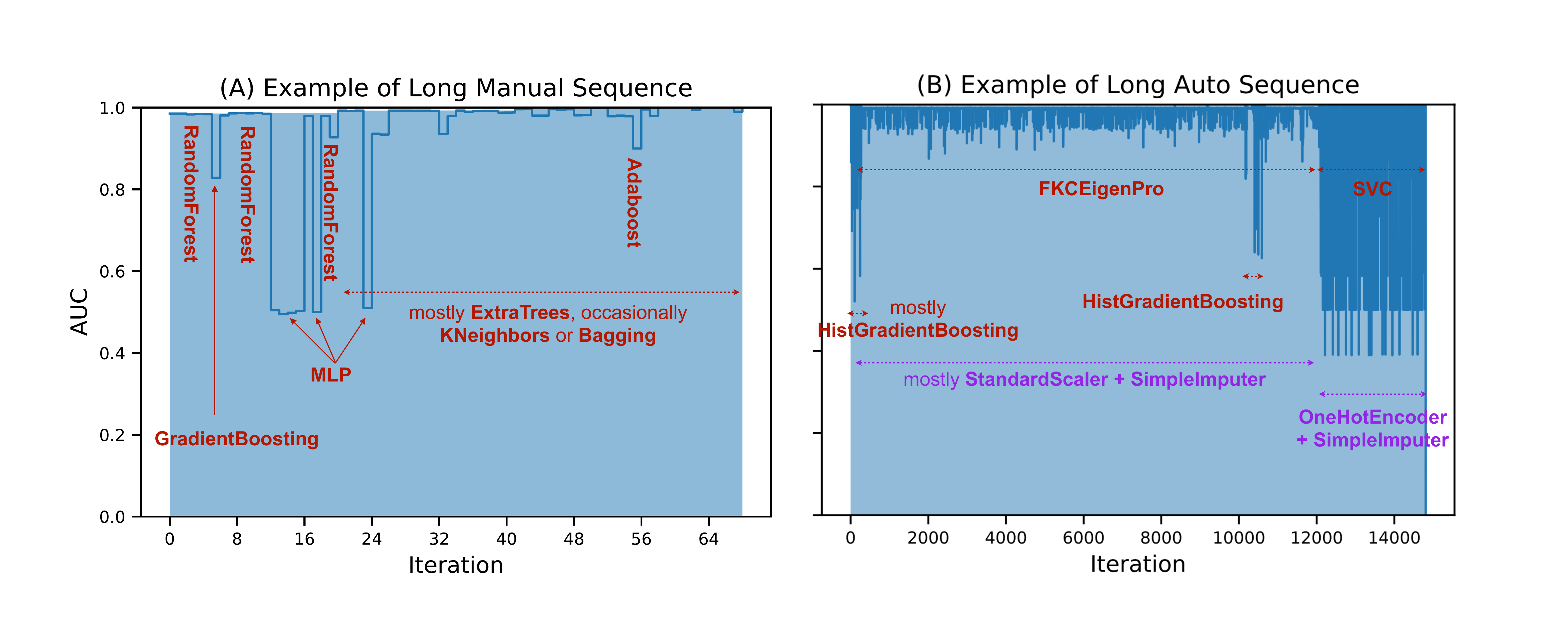}%
\caption{Longest manual sequence (A) and longest auto sequence (B). The shaded blue area represents the maximum AUC up until that point, the red text describes model changes, and the purple text describes preprocessing changes.}
\label{fig:casestudieslongest}
\end{figure*}

Two additional interesting examples are shown in Figure~\ref{fig:casestudieslongest}. The manual sequence with the highest number of iterations among all manual sequences is shown in (A) of the figure, while the longest auto sequence is shown in (B). The task in the manual sequence (A) 
was supervised classification on EEG data with 14 different EEG measurements for the features, with a binary label for whether the eye was closed or open (55.12\% closed)~\cite{Dua:2019}. The auto sequence (B) 
was performed for a supervised classification task on a chess dataset, with 36 board positions for the features, and a binary label for whether or not White can win over Black (52\% positive)~\cite{Dua:2019}.
In the auto example, the high fluctuation (dense zig-zagging pattern) in AUC over time seems to be due to random values being used for hyperparameters. While certain models (SVC in this instance) have a wide range of possible AUC scores, others (such as FKCEigenPro) have a much narrower range that AUC bounces around in. This kind of behavior is less visible in manual sequences, where the AUC range of each model is much smaller, while the biggest differences are a result of switching to a different model. For instance, from Figure~\ref{fig:casestudieslongest}(A), there is a noticeable drop in AUC when going from Random Forest to Gradient Boosting, and from Random Forest to MLP. However, within the Random Forest iterations themselves, we do not see high fluctuations, and this holds true for the MLP and ExtraTrees models as well.

These five examples in Figures~\ref{fig:casestudiestypes} and~\ref{fig:casestudieslongest} not only provide a glimpse of potential differences between iterative development styles and their resulting performance patterns, but also demonstrate the usefulness of tracking changes to an ML workflow and visualizing progress over time. If there were a real-time visualization tool that displayed the kind of information shown in these figures, this could potentially help users catch ineffective patterns earlier on in their iterations. For instance, perhaps if the user in Figures~\ref{fig:casestudiestypes}(C) were to see the stairstep-like subpattern occurring two or three times in a row, they would have realized that incrementing the number of neighbors in KNeighbors followed by the number of components in PCA is an ineffective strategy in raising the maximum AUC.

Next, we will discuss specific indicators of effective and ineffective workflow sequences. In order to make fair comparisons across sequences, we choose from among the most effective and least effective of sequences belonging to the same task.

\subsection{On Performance on the Same Task}
To learn what sets apart sequences that are highly effective and ones that are ineffective, we first define a highly effective sequence as one that achieves a high AUC score over just a few iterations, wasting few to no iterations after the maximum AUC has been achieved. Meanwhile, ineffective sequences exhibit opposite characteristics, namely, they span many more iterations but do not achieve high AUC scores.

\balance
\begin{figure*}
  \includegraphics[clip,width=\textwidth]{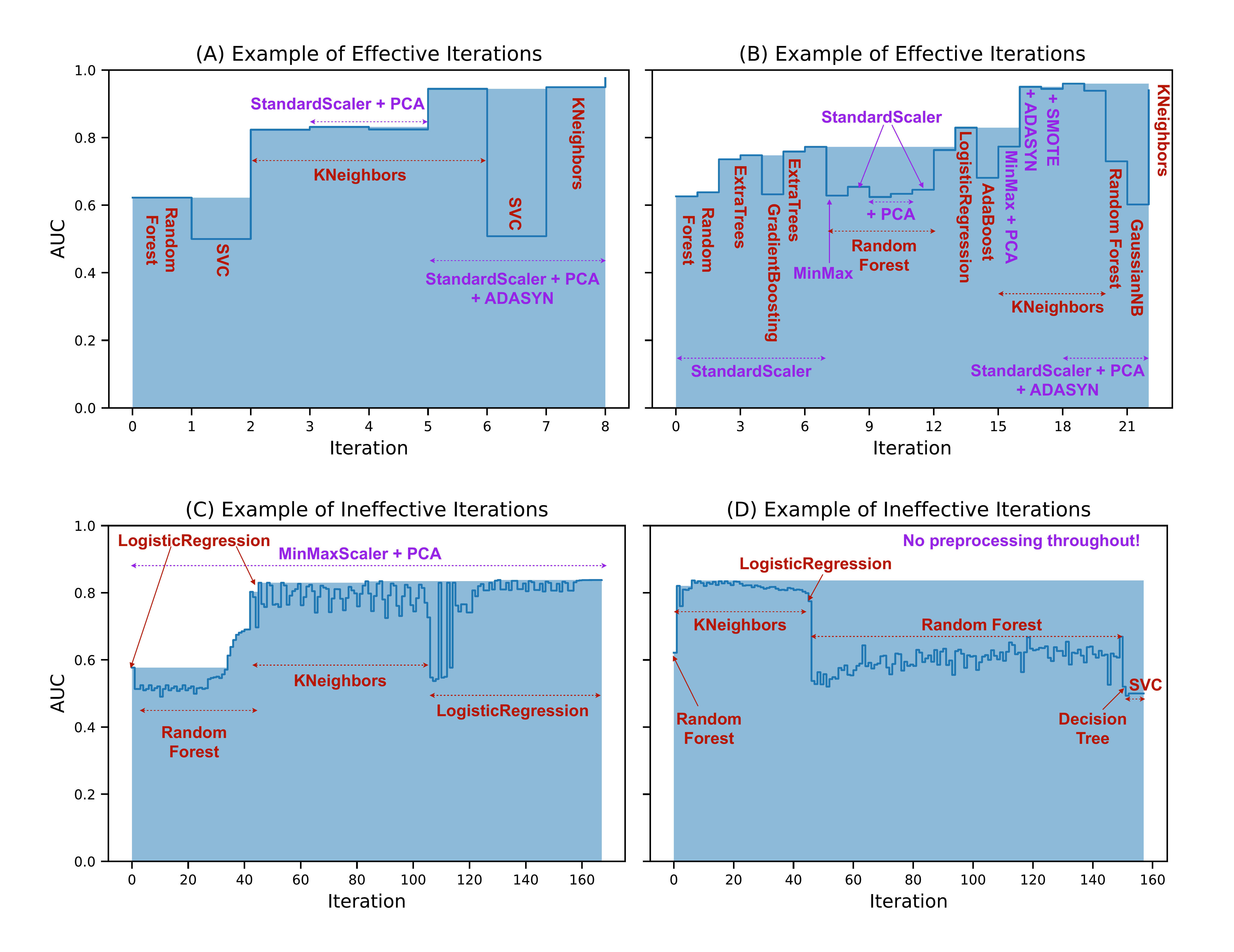}%
\caption{Examples of effective (A, B) and ineffective (C, D) sequences for a popular classification task. The shaded blue area represents the maximum AUC up until that point, the red text describes model changes, and the purple text describes preprocessing changes.}
\label{fig:casestudiesperf}
\end{figure*}

In Figure~\ref{fig:casestudiesperf}, we show examples of  highly-effective sequences (A 
and B) 
and ineffective sequences (C 
and D) 
for the supervised classification task that was attempted by the highest number of users on OpenML. The speech dataset used in this task has 3,686 instances, 400 features, and 2 classes: whether the speaker (row) had an American accent or not~\cite{speechdataset}. Most notably, there is a very imbalanced class distribution, with 98.35\% of the instances belonging to the majority class (positive for American accent). Case studies for this task demonstrate that workflows that account for the class imbalance problem are able to achieve higher performance than those who do not. 

Figure~\ref{fig:casestudiesperf}(A) illustrates one such example of a user who was successful in creating the best ML workflow for this task, with a higher maximum AUC score than any other user. This expert user (who will we refer to as User A) iterated on the workflow using a \emph{manual} sequence (based on our definition in Section~\ref{sec:seq}), starting off by selecting a model, then adding data preprocessing, beginning with StandardScaler and Principal Component Analysis (PCA). User A then incorporated an oversampling strategy, ADASYN, to counter the the class imbalance problem, resulting in a significant performance boost. 
Perhaps to validate that KNeighbors in conjunction with the 3 preprocessing operators are contributing to the high AUC, rather than solely the preprocessing operators, User A switched back to SVC at iteration 6, resulting in a large drop in performance. In the next iteration, they quickly returned to KNeighbors, demonstrating strong iterative development skills due to the fact that the user was able to quickly verify a hypothesis without wasting more than one iteration; they were able to recognize the exact operator combinations that worked well and move forward accordingly.

Similarly, Figure~\ref{fig:casestudiesperf}(B) illustrates another user who iterated with a \emph{manual} sequence and experimented with different combinations of model and preprocessing operators before discovering the high impact of data oversampling. It took this user more iterations than User A, but as soon as they incorporated ADASYN, they too were able to quickly run through a final model selection phase to achieve one of the highest AUC scores for the task. In both of these examples of effective sequences of workflow changes, the evolution of the preprocessing steps included in the workflow show that the user was actively reasoning about whether or not certain kinds of data operations were needed as they adjusted the workflow. However, not all users put as much thought into deciding on a strong data preprocessing combination.

For instance, User C (shown in Figure~\ref{fig:casestudiesperf}(C)) had an ineffective sequence of mostly \emph{automated} hyperparameter (and preprocessing hyperparameter parameter) changes. Extensively fine-tuning model and preprocessing hyperparameters might have been useful once a high-potential model and preprocessing combination was determined. But because the tuning was done prematurely to finding a solution to handle the imbalanced classes, User C was unable to develop a workflow that performed as well as the highly effective users (Users A and B). This user would have benefited from experimenting with other preprocessing combinations rather than completing all 168 iterations with the same set of preprocessing operators (MinMaxScaler and PCA). However, still other users were excluding preprocessing from their workflows completely.

As shown in Figure ~\ref{fig:casestudiesperf}(D), the user was seemingly off to a strong start, having chosen the same model that Users A and B began with, then quickly moving on to KNeighbors, which was the model that the two experts ended up selecting.
However, User D then spent over one hundred \textit{automated} iterations exhaustively tuning some hyperparameters, resulting in over 95\% of their iterations being wasted, i.e., they did not improve their maximum AUC. In both examples of ineffective iterations, the variety of model and preprocessing combinations covered was relatively limited (or nonexistent in the case of User D's lack of preprocessing), especially given the number of iterations spent on developing the workflow. Their generally ineffective pattern appears to be tuning each model for a few dozen iterations before trying a different model and possibly revisiting old models.
Users C and D could benefit from learning the strategy adopted in manual approaches in Figure~\ref{fig:casestudiesperf}(A) and (B) to first examine dataset characteristics to guide modeling decisions, instead of optimizing hyperparameters prematurely. 

These case studies demonstrate the potential benefits of knowledge transfer from experts to novices, conceivably through a human-in-the-loop system for ML workflow development using crowdsourced best practices to guide users towards constructing optimal workflows.

\section{Discussion and Conclusion}
\label{sec:conclusion}
In this paper, we set out to demystify the dark art of machine learning workflow development. Our analyses spanned multiple different levels---from aggregate statistics at the run-level, using over 475 thousand user-generated runs on OpenML; to aggregate statistics at the sequence-level, using 2557 sequences of runs (grouped by user and task); to individual case studies of sequences to learn from concrete real-world examples. 
We discovered that the performance improvement gap between manual and automated approaches to workflow iteration is negligible, but the manual approach is able to find the best performing workflow with much fewer wasted iterations. 
The model performance parity explains the general enthusiasm around automated machine learning (auto-ML) systems in recent years, but the massive discrepancy in efficiency explains the sluggish growth in adoption of auto-ML despite the hype. 
Our case studies revealed how human users are often far better at minimizing wasted work than non-optimized automation.

While we believe that the insights we have compiled are generalizable to the larger machine learning community, there is a chance that the conclusions we have drawn are applicable to only the population of users on the OpenML platform. We leave it to future work to perform similar analyses on ML workflow data from other platforms such as D3M~\cite{d3m} or Kaggle\footnote{https://www.kaggle.com/}. If the insights vary significantly between the different platforms, it would be interesting to further investigate the cause of these differences and learn what factors are driving certain kinds of ML developers into using certain platforms.

Another potential extension of this work would be to examine even more finer-grained categories of workflow changes. For instance, rather than grouping all kinds of model hyperparameter changes together, we could distinguish between the \emph{number} of hyperparameters that were changed from one iteration to the next (so changing only one hyperparameter would be a different category than changing three). The drawback to these finer-grained analyses is that the more detailed the categorization, the more sparse the samples that fall into each category, resulting in weaker (less generalizable) aggregate findings. Many of the decisions that we made in designing our analyses required such balancing acts, but because we have not exhaustively tried all possibilities, we cannot claim to have reached the optimal categorizations and thresholds. However, we have made a great effort to boost the integrity and robustness of our insights through our multi-faceted data summarizations and multiple levels of granularity as mentioned earlier.

In addition, while our study surfaced many interesting patterns in iterative ML workflow development, our dataset provides a limited view into the user's thought process behind performing the manual iterations, as shown in the case studies. 
A more in-depth user study is a promising direction for future work towards understanding the motivations and cognitive processes behind \emph{how} and \emph{why} users select certain iterative workflow changes. 
This understanding can form the basis of an improved auto-ML strategy that is drastically more efficient by mimicking human experts, thus lowering the barriers to entry to auto-ML adoption.
Likewise, human-in-the-loop ML systems can similarly benefit from automated guidance suggesting areas of exploration
that the ML developer may not have thought of. 
Our study aims to shed light on design insights for building
more usable and more intelligent machine learning development systems, towards the ultimate goal of democratizing machine learning.

\smallskip
\noindent {\bf Acknowledgments.} We thank the anonymous reviewers for their valuable feedback. We acknowledge support from grants IIS-1652750 and IIS-1733878 awarded by the National Science Foundation, grant W911NF-18-1-0335 awarded by the Army, and funds from the Alfred P. Sloan Foundation, Facebook, Adobe, Toyota Research Institute, Google, and the Siebel Energy Institute. The content is solely the responsibility of the authors and does not necessarily represent the official views of the funding agencies and organizations.

\bibliographystyle{ACM-Reference-Format}
\bibliography{sources}

\end{document}